\documentclass[pdflatex,sn-nature,onecolumn,oneside]{sn-jnl}

\usepackage[T1]{fontenc}
\usepackage{times}
\usepackage{graphicx}
\usepackage{amsmath,amssymb,amsfonts}
\usepackage{textcomp}
\usepackage{booktabs}
\usepackage{tabularx}
\usepackage{array}
\usepackage{etoolbox}
\usepackage{balance}
\usepackage{xcolor}
\usepackage[most]{tcolorbox}
\usepackage{helvet}
\usepackage{microtype}

\usepackage{colortbl}
\definecolor{TableRule}{HTML}{D2D5DA}
\newcommand{\tablerowrule}{%
  \arrayrulecolor{TableRule}%
  \specialrule{0.25pt}{2.2pt}{2.2pt}}

  \newcommand{\densetablerowrule}{%
  \arrayrulecolor{TableRule}%
  \specialrule{0.25pt}{1.5pt}{1.5pt}}

\definecolor{FulliveLavender}{HTML}{f2f4f9}
\definecolor{FullivePurple}{HTML}{4e588f}
\definecolor{FulliveInk}{HTML}{1C2B33}

\newcommand{\FulliveLogo}[1]{\includegraphics[height=#1]{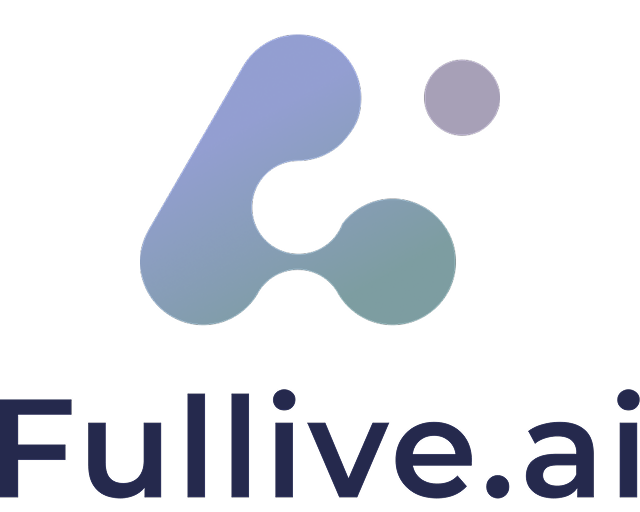}}
\newcommand{\PekingUniversityLogo}[1]{\includegraphics[height=#1]{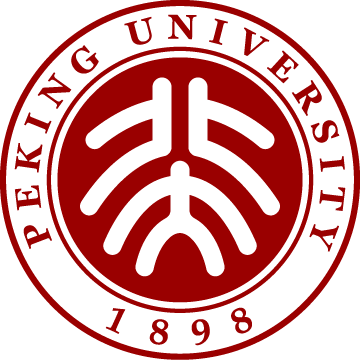}}

\newcommand{\InstitutionLogoLockup}{%
  \FulliveLogo{0.38in}%
  \hspace{0.08in}%
  \raisebox{0.10in}{\rule{0.4pt}{0.14in}}%
  \hspace{0.10in}%
  \PekingUniversityLogo{0.34in}%
}

\newcommand{\ArticleTypeLabel}{%
  \sffamily\bfseries
  \fontsize{14}{14}\selectfont
  \color{FulliveInk}
  Perspective%
}

\makeatletter
\def\Titlefont{\reset@font\sffamily\fontsize{16.2}{19.1}\bfseries\selectfont\raggedright}
\def\Authorfont{\reset@font\sffamily\fontsize{10.2}{12.4}\bfseries\color{FullivePurple}\selectfont\raggedright}
\def\addressfont{\reset@font\rmfamily\fontsize{9.8}{11.8}\color{FullivePurple}\selectfont\raggedright}
\def\authornotefont{\reset@font\rmfamily\fontsize{9.4}{11.2}\color{FullivePurple}\selectfont\raggedright}
\def\correspondencefont{\reset@font\rmfamily\fontsize{9}{10.8}\color{FulliveInk}\selectfont\raggedright}
\def\abstractfont{\reset@font\rmfamily\fontsize{10}{11.7}\color{FulliveInk}\selectfont\leftskip=0pt\rightskip=0pt}
\def\keywordfont{\reset@font\rmfamily\fontsize{10}{11.7}\color{FulliveInk}\selectfont}
\def\sectionfont{\reset@font\sffamily\fontsize{12.6}{15}\bfseries\selectfont\raggedright\boldmath}
\def\subsectionfont{\reset@font\sffamily\fontsize{10.5}{12.5}\bfseries\selectfont\raggedright\boldmath}
\def\subsubsectionfont{\reset@font\sffamily\fontsize{9.8}{11.8}\bfseries\selectfont\raggedright\boldmath}
\def\paragraphfont{\reset@font\sffamily\fontsize{9.8}{11.8}\bfseries\selectfont\raggedright\boldmath}
\def\bmheadfont{\reset@font\sffamily\fontsize{9.8}{11.8}\bfseries\selectfont\raggedright\boldmath}
\def\figurecaptionfont{\reset@font\fontsize{9}{10.5}\selectfont\raggedright}
\def\tablecaptionfont{\reset@font\fontsize{9}{10.5}\selectfont\raggedright}
\def\bibfont{\reset@font\fontsize{9}{10.5}\selectfont}

\renewcommand\section{\@startsection{section}{1}{\z@}%
  {-14pt \@plus -3pt \@minus -2pt}{7pt}{\sectionfont}}
\renewcommand\subsection{\@startsection{subsection}{2}{\z@}%
  {-10pt \@plus -2pt \@minus -1pt}{4pt}{\subsectionfont}}
\renewcommand\subsubsection{\@startsection{subsubsection}{3}{\z@}%
  {-8pt \@plus -2pt \@minus -1pt}{3pt}{\subsubsectionfont}}
\renewcommand\paragraph{\@startsection{paragraph}{4}{\z@}%
  {7pt \@plus 1pt \@minus 1pt}{-0.6em}{\paragraphfont}}
\def\@seccntformat#1{\csname the#1\endcsname\hskip0.75em}

\renewcommand\figurename{Figure}
\def\fnum@figure{\figurename\space\thefigure}
\def\fnum@table{\tablename\space\thetable}
\long\def\@figurecaption#1#2{{\figurecaptionfont\textbf{#1}\hskip0.5em#2\par}}
\long\def\@tablecaption#1#2{%
  \setbox\tabcapbox\vbox{\tablecaptionfont\raggedright%
  \textbf{#1}\hskip0.5em#2\vphantom{y}\par}%
  \box\tabcapbox}
\def\au@and{\unskip,\hspace{0.45em}}

\gdef\FulliveAbstract{}
\long\def\abstract#1{\gdef\FulliveAbstract{#1}}
\gdef\FulliveKeywords{}
\long\def\keywords#1{\gdef\FulliveKeywords{#1}}

\def\ps@fullivebody{%
  \def\@oddhead{}%
  \let\@evenhead\@oddhead
  \def\@oddfoot{\hfil\rmfamily\fontsize{9}{11}\selectfont\thepage\hfil}%
  \let\@evenfoot\@oddfoot}
\def\ps@fullivefirst{%
  \def\@oddhead{%
    \vbox to 30pt{%
      \hbox to\textwidth{%
        \raisebox{6pt}{\ArticleTypeLabel}%
        \hfill
        \raisebox{5pt}[0.40in][0pt]{\InstitutionLogoLockup}%
      }%
      \vfill
      \hrule height0.6pt width\textwidth
    }%
  }%
  \let\@evenhead\@oddhead
  \def\@oddfoot{}%
  \let\@evenfoot\@oddfoot
}

\renewcommand\maketitle{%
  \thispagestyle{fullivefirst}%
  \vspace*{0pt}%
  \begin{tcolorbox}[
    enhanced,colback=FulliveLavender,colframe=FulliveLavender,
    boxrule=0pt,arc=10pt,outer arc=10pt,boxsep=0pt,
    left=20pt,right=20pt,top=16pt,bottom=14pt,
    before skip=0pt,after skip=15pt,width=\textwidth]
    {\Titlefont\begingroup\let\vspace\@gobble\@title\endgroup\par}
    \vspace{7pt}
    {\artauthors\par}

    \ifx\auaddress\@empty\else
      \vspace{5pt}
      {\addressfont
        \begingroup
        \def\par{\unskip\hspace{1.4em}}%
        \auaddress
        \endgroup\par}
    \fi

    \ifequalcont
      \vspace{2pt}
      {\authornotefont
        $^{\dagger}$Equal contribution,\hspace{1em}%
        $^{*}$Corresponding author\par}
    \fi

    \vspace{14pt}
    {\abstractfont\FulliveAbstract\par}
    \vspace{9pt}
    {\keywordfont\raggedright\textbf{Keywords:} \FulliveKeywords\par}
  \end{tcolorbox}%
  \pagestyle{fullivebody}%
}
\makeatother

\hypersetup{colorlinks=true,linkcolor=FullivePurple,citecolor=FullivePurple,urlcolor=FullivePurple}

\begin{document}

\title[Physiological World Models]{Physiological world models for human state transitions\vspace{13pt}}

\author[1]{\fnm{Chongyang} \sur{Zhang}}
\equalcont{These authors contributed equally to this work.}

\author[1]{\fnm{Rendong} \sur{Wang}}
\equalcont{These authors contributed equally to this work.}

\author[1]{\fnm{Hao} \sur{Zheng}}

\author[1]{\fnm{Hanwen} \sur{Zhang}}

\author[2]{\fnm{Yang} \sur{Liu}}

\author[1]{\fnm{Xiaolong} \sur{Wei}}

\author*[2]{\fnm{Bin} \sur{Chong}}

\affil[1]{\orgname{Fullive-AI}}

\affil[2]{\orgname{Peking University}}

\abstract{Continuous multimodal sensing now allows human physiology to be observed throughout daily life rather than only during occasional clinical visits. However, most health artificial intelligence systems are designed to recognize current states, estimate risks or analyse individual biomarkers. They do not directly model how physiological states change in response to real-world events, behaviours, contexts and interventions. Here we propose the Physiological World Model (PWM), an event-conditioned framework for learning these changes at the level of the whole person. We introduce the HumanState Transition Token, a structured, quality-scored unit that connects the physiological state before an event with the event or action, relevant context and intervention information, the physiological trajectory after the event, observed outcomes and data quality. We describe four capability levels, from state representation to bounded intervention planning, together with four data acquisition and validation protocols. We also propose six benchmark tasks covering HumanState representation, forecasting across multiple timescales, individualized response prediction, simulation of alternative interventions, bounded planning and reliability under distribution shift. Together, this framework provides a practical path towards personalized health management, behavioural intervention design and clinician-supervised decision support, while clearly separating prediction from causal inference and making uncertainty, safety, governance and limits of use explicit.}

\keywords{Physiological World Models, Event-conditioned Modeling, Causal Inference, Counterfactual Simulation, Synchronized Multimodal Data Protocol, Digital Health}

\maketitle

\setlength{\abovedisplayskip}{14pt plus 2pt minus 1pt}
\setlength{\belowdisplayskip}{14pt plus 2pt minus 1pt}
\setlength{\abovedisplayshortskip}{14pt plus 2pt minus 1pt}
\setlength{\belowdisplayshortskip}{14pt plus 2pt minus 1pt}

\section{Introduction}\label{sec:introduction}

Observing human physiology continuously in the real world, rather than episodically in the clinic, represents one of the most consequential shifts in personal health monitoring since the invention of the Holter monitor~\cite{ref110}. Over the past decade, the integration of miniaturized sensors, ubiquitous wireless connectivity, and cloud-scale data infrastructure has brought multimodal physiological monitoring to daily life on an unprecedented scale~\cite{ref110,ref111}. Smartwatches and fitness bands capture cardiovascular, activity, temperature, and electrodermal signals while estimating sleep patterns~\cite{ref53,ref101,ref111,ref112}; continuous glucose monitors and wearable ECGs extend metabolic and cardiac monitoring beyond the clinic~\cite{ref12,ref113,ref114}; and smartphones with environmental sensors record contextual factors such as light, noise, temperature, and location, which are increasingly integrated with physiological time series~\cite{ref105,ref106,ref115,ref116}. However, the infrastructure for modelling and reasoning over these data has not kept pace with data collection~\cite{ref117,ref118}. Physiological records are generated and organized around devices, care settings, disease categories, and institutional silos, and are typically treated as modality- or task-specific features rather than as coordinated representations of an evolving human state~\cite{ref102,ref103}. As a result, real-world events, individual behaviours, physiological responses, and intervention outcomes are rarely aligned along a common temporal axis. A unified modelling framework is therefore needed to integrate multimodal observations and characterize how an individual's physiological state evolves in response to specific events and actions in defined contexts~\cite{ref106,ref107,ref108,ref109}.

Artificial intelligence has been increasingly applied to health data. Yet most current systems are designed for state recognition, anomaly detection, or risk prediction. They answer what an individual's current physiological state is, whether a biomarker is abnormal, or whether a future clinical outcome is likely~\cite{ref8,ref13,ref14,ref119,ref120}. Although these functions have demonstrated clinical utility~\cite{ref114,ref119}, current systems often model physiology through isolated observations or input--output associations, rather than as a dynamic process shaped by events, behaviours, and context~\cite{ref118,ref121}. What remains insufficiently addressed is how physiological states change over time: why a state changes, why the same event produces different responses across individuals, and how a trajectory might differ under alternative actions. Addressing this gap requires models that learn event-conditioned physiological state transitions, rather than merely recognizing current states or predicting predefined outcomes~\cite{ref6,ref117,ref118,ref121}.

Recent advances in world modelling provide a conceptual basis for extending world-model principles to human physiology. World models learn task-relevant state representations and transition dynamics that support prediction, simulation, and planning~\cite{ref1,ref2,ref121}. They have advanced game-playing agents, robotic control, and autonomous navigation~\cite{ref3,ref4,ref100,ref121}, and related approaches are beginning to appear in biomedical research, including molecular and cellular modelling, physiological-signal modelling, and clinical trajectory analysis~\cite{ref5,ref6,ref7,ref36,ref38,ref39}. However, existing biomedical world-model research has rarely centred on how an individual's macro-level physiology evolves in response to real-world events and interventions. Modelling physiology at this scale requires temporally aligning multimodal physiological observations with real-world events, actions, interventions, and contexts across multiple timescales~\cite{ref112,ref115,ref116}.

\begin{figure*}[t]
\centering
\includegraphics[width=\textwidth]{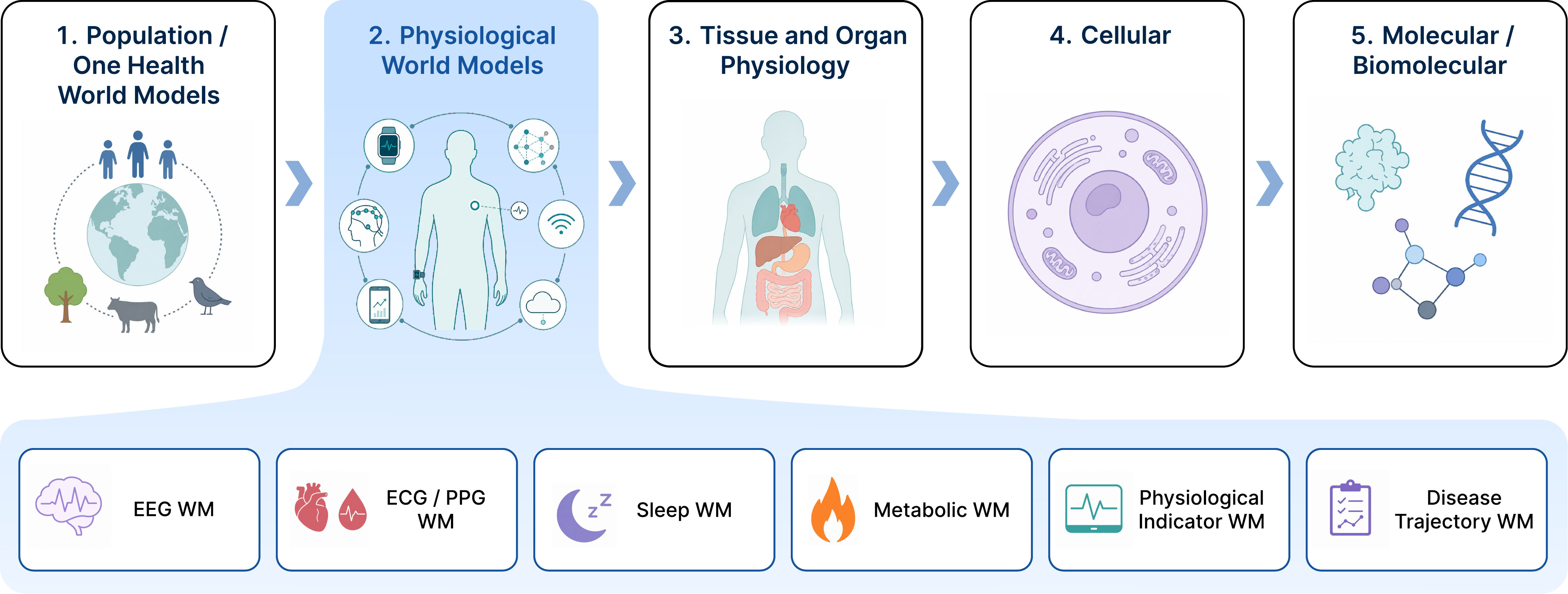}
\caption{\textbf{Physiological world models within a multiscale hierarchy of biological and health modelling.} The proposed taxonomy orders modelling scopes from population and One Health systems~\cite{ref175}, through integrated physiology at the individual level, to tissue and organ~\cite{ref26}, cellular~\cite{ref173}, and molecular or biomolecular processes~\cite{ref176}. PWMs occupy the individual-level layer, whereas models at broader scales address population, cross-species, and environmental dynamics, and models at finer scales resolve processes within organs, cells, or molecules. The lower panel shows illustrative domain-specific specializations within the PWM level, including EEG, ECG/PPG~\cite{ref172}, sleep, metabolic, physiological-indicator, and disease-trajectory WMs~\cite{ref25,ref38,ref39}. These subdomains include both existing model instances and prospective directions. Arrows indicate the ordering of modelling scopes rather than causal or computational flow.}\label{fig:world-model-scales}
\end{figure*}

In this Perspective, we define the Physiological World Model (PWM) as an event-conditioned framework for modelling transitions in an individual's macro-level physiological state. Whereas digital twins generally emphasize individual-specific virtual representations and continuous state synchronization~\cite{ref75,ref76}, PWMs focus on learning event-conditioned physiological state-transition dynamics; we discuss this distinction further in Section~\ref{sec:scope-boundaries-and-relation-to-digital-twins}. As illustrated in Figure~\ref{fig:world-model-scales}, our proposed taxonomy positions PWMs at the level of integrated individual physiology, between models of broader population and One Health systems and finer-scale models of tissues, organs, cells, and molecules~\cite{ref175, ref26, ref173, ref151}. To support the modelling of physiological state transitions, we introduce the HumanState Transition Token as a structured, quality-scored unit for transition learning and evaluation. We then describe an architecture-agnostic learning framework and an L1--L4 PWM capability ladder extending from state representation to bounded intervention planning. Four levels of data acquisition and validation (P1--P4) provide progressively stronger forms of evidence for these capabilities, although the correspondence is not one-to-one. We further propose six complementary benchmark tasks covering HumanState representation, multi-horizon forecasting, event-conditioned and individualized response prediction, alternative intervention simulation, bounded planning, and reliability under distribution shift. These tasks are linked to near-term personal health applications, mid-term behavioural intervention design, and longer-term clinical decision support, with progressively stronger validation and governance requirements~\cite{ref66,ref67,ref81,ref119,ref120,ref121}.

\section{Defining Physiological World Models}\label{sec:defining-physiological-world-models}

\subsection{Background and Scope}\label{sec:background-and-scope}

World models have recently emerged as a class of approaches for learning predictive representations of dynamical systems in artificial intelligence and reinforcement learning~\cite{ref1,ref2,ref3,ref4}. Rather than reconstructing an environment in full, they learn task-relevant state representations and transition dynamics that support forecasting, simulation, and planning~\cite{ref1,ref3,ref4,ref55}. World-model approaches have begun to appear in clinical trajectory modelling and decision support~\cite{ref6}, while related predictive-representation methods are being explored in wearable biosignal modelling~\cite{ref7}. However, approaches that connect real-world events, human actions, contexts, and interventions to transitions in an individual's integrated physiological state remain underexplored.

From this perspective, we define the Physiological World Model (PWM) as a human-centred, event-conditioned model of the transition dynamics of an individual's integrated physiological state. Here, human-centred means that the evolving physiological state of the whole person, situated within their physical, social and behavioural environment, is the primary object of modelling. A PWM therefore focuses on macro-scale, temporally evolving physiology rather than attempting to reconstruct human biology in its entirety or being limited to episodic clinical records. Its distinguishing objective is to connect internal physiological dynamics with external events, actions, contexts, and interventions. This enables future state trajectories to be predicted and, when supported by appropriate interventional evidence, compared across alternative actions or interventions.

A PWM does not treat physiological and contextual data as a flat collection of features. Instead, it represents physiological information across three levels while retaining contextual measurements as distinct conditioning variables. The Raw Observation Layer contains physiological sensor streams and time-aligned contextual measurements. The Interpretable Physiology Layer transforms physiological observations into interpretable intermediate variables, such as autonomic regulation, metabolic stability, circadian phase, sleep pressure, and recovery capacity~\cite{ref11,ref12,ref90,ref95,ref160}. The Latent HumanState Layer integrates these physiological observations and intermediate variables into an individual's macro-level physiological state.

Together, these layers connect heterogeneous measurements to a unified state representation without assuming that all devices measure the same variables or provide the same level of fidelity. The resulting latent state supports transition prediction, simulation, and intervention comparison. Having defined what a PWM represents, we next formalize how that state changes over time.

\subsection{HumanState and Transition Dynamics}\label{sec:human-state-and-transition-dynamics}

Beyond representing an individual's current physiological state, a PWM models how that state evolves in response to events, actions and interventions within defined contexts. Standard stochastic world models commonly represent action-conditioned dynamics through a transition distribution of the form
\(p(s_{t+1}\mid s_t,a_t)\)~\cite{ref121}. We extend this formulation by incorporating observed context and an explicit prediction horizon:
\begin{equation}
p\!\bigl(
s_{t+\Delta t}
\mid
s_t,a_t,c_t,\Delta t
\bigr).
\label{eq:generic-transition}
\end{equation}
\noindent Here, \(s_{t}\) and \(s_{t + \Delta t}\) denote the current and future states, respectively; \(a_{t}\) denotes an action; \(c_{t}\) denotes contextual information; \(t\) denotes the current time; and \(\Delta t\) denotes the prediction horizon. Extending this formulation to human physiology, a PWM models the conditional distribution of future HumanState as:
\begin{equation}
\begin{aligned}
p\!\bigl(\mathrm{HumanState}_{t+\Delta t}\mid {}&
\mathrm{HumanState}_t,\mathrm{WorldEvent}_t,\mathrm{HumanAction}_t,\mathrm{Context}_t,\mathrm{Intervention}_t,\Delta t\bigr).
\end{aligned}
\label{eq:pwm-transition}
\end{equation}
\noindent This formulation aligns the PWM with a general world-model transition formulation while making the drivers of physiological change explicit. It allows the model to represent multiple plausible future trajectories and their associated uncertainty via multi-step rollouts, rather than returning only a deterministic point estimate. Comparisons across alternative interventions should be interpreted as causal effects only when supported by appropriate study designs and assumptions. This formulation specifies the conceptual dependencies governing physiological state transitions, without prescribing how the variables are encoded or which neural architecture is used to model them. The terms in the PWM formulation are defined as follows.

\(\mathrm{HumanState}_{t}\) denotes the latent macro-level physiological state of an individual at time \(t\). It is inferred from an integration of physiological observations such as heart rate, blood glucose, blood pressure, oxygen saturation, body temperature, and validated sleep-related measures with relevant disease-related status, including chronic conditions, comorbidities, and disease stage, where available. Rather than corresponding to any single observed measure, \(\mathrm{HumanState}_{t}\) reflects the individual's overall physiological condition and the accumulated effects of prior physiological changes, behaviours, and treatments up to time \(t\). Where reliable assessments are available, subjective variables such as mood and perceived stress may also inform the inference of this state~\cite{ref15}.

\(\mathrm{WorldEvent}_{t}\) captures external events and exposures that may influence HumanState. Examples include abrupt changes in environmental noise, ambient temperature, or light exposure; traffic disruptions; salient social interactions; and unexpected work-related demands. \(\mathrm{HumanAction}_{t}\) represents observed volitional behaviours, such as eating, exercise, caffeine or alcohol intake, medication adherence, screen use, and sleep timing.

\(\mathrm{Context}_{t}\) specifies the observed individual and situational information available at or before time \(t\) that conditions a state transition but is not itself represented as the contemporaneous HumanState, focal WorldEvent, HumanAction, or Intervention. It may include time-invariant or slowly varying individual background, such as age, body size, past medical history, preferences, and habitual behaviours, as well as time-varying pre-transition context, such as time of day, day of week, previous-night sleep quality, and recent behavioural or dietary history. Actual eating, exercise, and other focal behaviours should be represented as HumanAction, whereas current physiological measures and current disease status belong to HumanState.

\(\mathrm{Intervention}_{t}\) refers to a deliberately assigned or planned modification intended to alter a physiological trajectory. Examples include a dietary regimen, exercise prescription, sleep-schedule adjustment, medication change, or behavioural modification programme. This definition distinguishes a planned intervention from a naturally occurring or merely observed action, even when the two involve the same behaviour.

Finally, \(\mathrm{HumanState}_{t + \Delta t}\) represents the future physiological state at the specified prediction horizon \(\Delta t\). A physiological trajectory can be obtained by rolling out the transition model across multiple time steps. The horizon may range from minutes to years, depending on the task. Postprandial glucose responses and heart-rate recovery require short-horizon prediction, whereas physiological adaptations to sustained training, chronic disease progression, and other long-term physiological changes require substantially longer horizons~\cite{ref89,ref161}. Different horizons may also require different temporal resolutions or hierarchical transition components.

\subsection{HumanState Transition Token}\label{sec:humanstate-transition-token}

Continuous recordings provide the observational foundation for PWM development, but transition learning requires event-anchored samples constructed from these streams. The state transition is therefore the fundamental modelling object of a PWM, while a HumanState Transition Token is the corresponding structured data instance for training and evaluation:
\begin{equation}
\begin{aligned}
\mathrm{HumanStateTransitionToken}
={}&\bigl(
\mathrm{HumanState}_{t},\,
\mathrm{WorldEvent}_{t},\,
\mathrm{HumanAction}_{t},\,
\mathrm{Context}_{t},\\
&\mathrm{Intervention}_{t},\,
\mathrm{HumanState}_{t+\Delta t},\,
\mathrm{Outcome}_{t+\Delta t},\\
&\mathrm{DataQuality}_{t:t+\Delta t}
\bigr).
\end{aligned}
\label{eq:transition-token}
\end{equation}
\noindent A transition token is not an arbitrary time window extracted from a continuous recording. It is anchored to an identifiable event, behaviour, or intervention around which a physiological response can be evaluated. Each token includes a pre-event baseline, a post-event response window, contextual variables, intervention information when applicable, and an observed outcome (for temporal descriptions such as pre-event and post-event windows, ``event'' denotes the focal transition anchor, which may be a WorldEvent, HumanAction, or Intervention). Fields that are not applicable to a given transition should be explicitly encoded as absent.

\noindent In Eq.~\eqref{eq:transition-token}, \(\mathrm{HumanState}_{t}\) and \(\mathrm{HumanState}_{t + \Delta t}\) denote inferred macro-level physiological states derived from the corresponding pre- and post-event observation windows, rather than directly observed measurements. By contrast, \(\mathrm{Outcome}_{t + \Delta t}\) denotes observable physiological readouts and task-specific endpoints at the evaluation horizon. Where an observation decoder is used, it maps the predicted state representation to \(\widehat{\mathrm{Outcome}}_{t + \Delta t}\), which can be compared with the observed \(\mathrm{Outcome}_{t + \Delta t}\) to assess the predicted transition. For example, pre-exercise heart rate, heart-rate variability, and glucose measurements may inform \(\mathrm{HumanState}_{t}\); post-exercise observations may inform \(\mathrm{HumanState}_{t + \Delta t}\), while the observed heart-rate recovery curve serves as an Outcome. \(\mathrm{DataQuality}_{t:t + \Delta t}\) summarizes the completeness, temporal alignment, sensor quality, annotation reliability, and overall confidence of the token across the full interval. These elements allow the reliability of each transition sample to be assessed explicitly.

This definition changes the scaling logic of PWM development. Traditional deep-learning approaches to physiological signals often treat scaling as the accumulation of larger volumes of continuous recordings~\cite{ref16,ref17}. For transition learning, however, the most informative samples are high-confidence HumanState Transition Tokens. A dataset may contain millions of hours of single-modality wearable data yet yield few usable transition tokens when event annotations and contextual labels are absent. By contrast, a smaller dataset may provide richer supervision when events, contexts, pre- and post-event physiological windows, interventions, and outcomes are carefully documented.

The role of transition tokens changes across capability levels. Large-scale unlabelled or weakly labelled recordings can support L1 state-representation learning. Progress from L2 state-transition prediction to L3 counterfactual simulation and L4 intervention planning increasingly depends on transition tokens with reliable event, action, context, intervention, outcome, and quality information. PWM development therefore requires not only more data, but also more diverse, higher-quality, and more information-dense transition evidence.

This scaling logic motivates the P1--P4 framework for data acquisition and validation. P1 characterizes baseline physiological distributions, P2 introduces structured event and context annotation, P3 adds assigned interventions and repeated observations, and P4 provides prospective shadow-mode validation and adaptive empirical feedback.

In summary, the basic unit for learning PWM transition dynamics is not a biomarker value at a single time point. It is a structured and verifiable record of how an individual's physiological state changes under specific events, actions, and interventions within defined contexts.

\section{Model Architecture and Learning Paradigm}\label{sec:model-architecture-and-learning-paradigm}

\begin{figure*}[t]
\centering
\includegraphics[width=\textwidth]{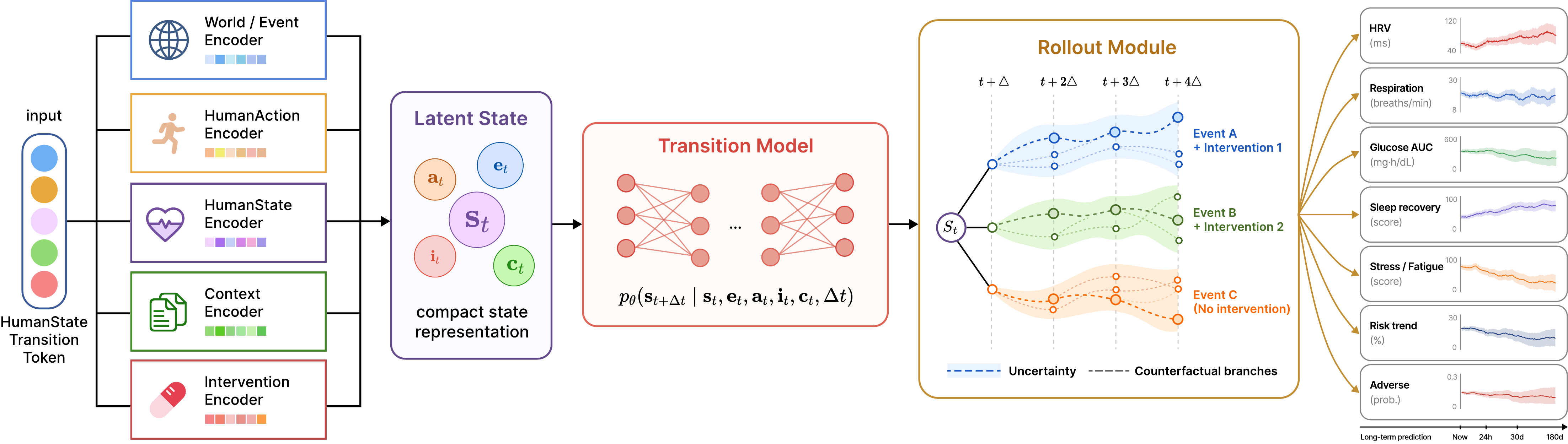}
\caption{\textbf{An illustrative functional architecture for a Physiological World Model.} Event, action, HumanState, context, and intervention are encoded through specific encoders and taken as conditional input to a latent-state transition model; rollout branches represent alternative future trajectories and uncertainty, and task-specific heads map predicted states to observable physiological outcomes.}\label{fig:pwm-architecture}
\end{figure*}

Section 2 defines what is represented by a PWM and the structure of HumanState Transition Tokens. Translating this definition into a trainable system requires both a latent state representation and a conditional transition model. Figure~\ref{fig:pwm-architecture} presents one possible functional architecture for such a system, informed by latent-dynamics and JEPA-style predictive learning~\cite{ref121,ref27,ref28}. In this illustrative architecture, the core components include a state encoder, a conditional transition model, an uncertainty-estimation component, and mechanisms for encoding events, actions, interventions, and context. Observation decoders or task-specific outcome heads may be added when required to map predicted state representations to observable outcomes. Counterfactual rollout and planning components are introduced at higher capability levels rather than treated as minimum requirements. These components can be implemented using different neural architectures. The schematic specifies functional requirements rather than prescribing a particular neural architecture.

The conditional transition distribution defined in Section 2 can be parameterized in latent space as:
\begin{equation}
p_{\theta}\!\bigl(\mathbf{s}_{t+\Delta t}\mid
\mathbf{s}_t,\mathbf{e}_t,\mathbf{a}_t,\mathbf{i}_t,\mathbf{c}_t,\Delta t\bigr),
\label{eq:latent-transition}
\end{equation}
\noindent where \(\mathbf{s}_{t}\) and \(\mathbf{s}_{t + \Delta t}\) are model-specific latent representations of \(\mathrm{HumanState}_{t}\) and \(\mathrm{HumanState}_{t + \Delta t}\), respectively, \(\mathbf{e}_{t}\) is an encoded representation of \(\mathrm{WorldEvent}_{t}\), \(\mathbf{a}_{t}\) is an encoded representation of \(\mathrm{HumanAction}_{t}\), \(\mathbf{i}_{t}\) is an encoded representation of a deliberately assigned or planned \(\mathrm{Intervention}_{t}\), and \(\mathbf{c}_{t}\) is an encoded representation of \(\mathrm{Context}_{t}\). The choice of horizon \(\Delta t\) depends on the physiological process, and different horizons may require different temporal resolutions or hierarchical transition components~\cite{ref6,ref21,ref84,ref85,ref86}.

Large-scale physiological observations can support state-representation learning, whereas HumanState Transition Tokens provide event-anchored supervision for learning conditional transition dynamics. Learning these representations and dynamics from real-world data presents several challenges: physiological observations are heterogeneous and irregularly sampled, events and interventions are sparsely and inconsistently labelled, and individual baselines vary substantially. The learning paradigms reviewed below offer complementary strategies for addressing these challenges.

\subsection{Learning Paradigms for PWMs}\label{sec:learning-paradigms-for-pwms}

Given the functional definition above, the remaining question is how to learn a robust and temporally coherent latent physiological state, together with its event-conditioned transition dynamics, from heterogeneous and partially observed data. Several complementary learning paradigms may contribute at different stages of PWM development, including model-based reinforcement learning, observation-level generative modelling, latent-dynamics learning, and JEPA-style predictive learning~\cite{ref6,ref26,ref27,ref28}.

Observation-level generative world models generate future trajectories in raw observation space, including medical images, ECG, PPG, sleep signals, or multimodal sequences~\cite{ref26,ref153,ref154}. They can learn complex temporal patterns, but these patterns do not necessarily correspond to intervention-relevant physiological states. Physiological data contain device noise, sampling variation, motion artefacts, and fluctuations unrelated to the individual's underlying physiological state. A reconstruction-heavy objective may therefore emphasize local morphology rather than underlying physiological dynamics~\cite{ref27,ref28}.

For this reason, we argue that latent-dynamics learning and JEPA-style prediction provide promising starting points for early-stage PWM development, potentially in combination with observation-level reconstruction objectives. These methods encode high-dimensional observations into latent states and predict future representations rather than every raw detail~\cite{ref27,ref28}. The model should preserve information relevant to physiological state and intervention, not every low-level fluctuation.

At later capability levels, model-based reinforcement learning provides an important conceptual reference for PWM development. Physiological condition can be represented as a latent state, while observed behaviours and assigned treatments can be represented as actions or interventions. Environmental changes may be represented as WorldEvents or contextual variables, and as Interventions when deliberately manipulated. The model can then learn transitions conditioned on these variables and simulate possible future trajectories~\cite{ref25}. However, health data are predominantly observational and offline. Real-world interaction is costly and ethically constrained; rewards are difficult to define; and the effects of observed actions are often confounded. Model-based reinforcement learning is therefore better viewed as a potential route towards L4 planning and closed-loop control rather than as the sole starting point for early-stage training~\cite{ref6,ref25,ref132,ref133,ref134}.

A practical development path can be organized into three stages. First, multimodal self-supervised learning, including JEPA-style objectives, is used to construct stable latent representations of HumanState. Building on these representations, HumanState Transition Tokens are then used to learn how physiological states evolve in response to specific events, actions, and interventions within defined contexts. Finally, the learned transition dynamics may support intervention comparison and planning, but only when high-confidence transition data, causally informative evidence, calibrated uncertainty, and explicit safety constraints are available. Physiology-informed constraints and prospective validation become increasingly important as the model advances towards counterfactual simulation and closed-loop use~\cite{ref6}.

\subsection{Capability Levels of PWMs}\label{sec:capability-levels-of-pwms}

\begin{figure*}[t]
\centering
\includegraphics[width=0.78\textwidth]{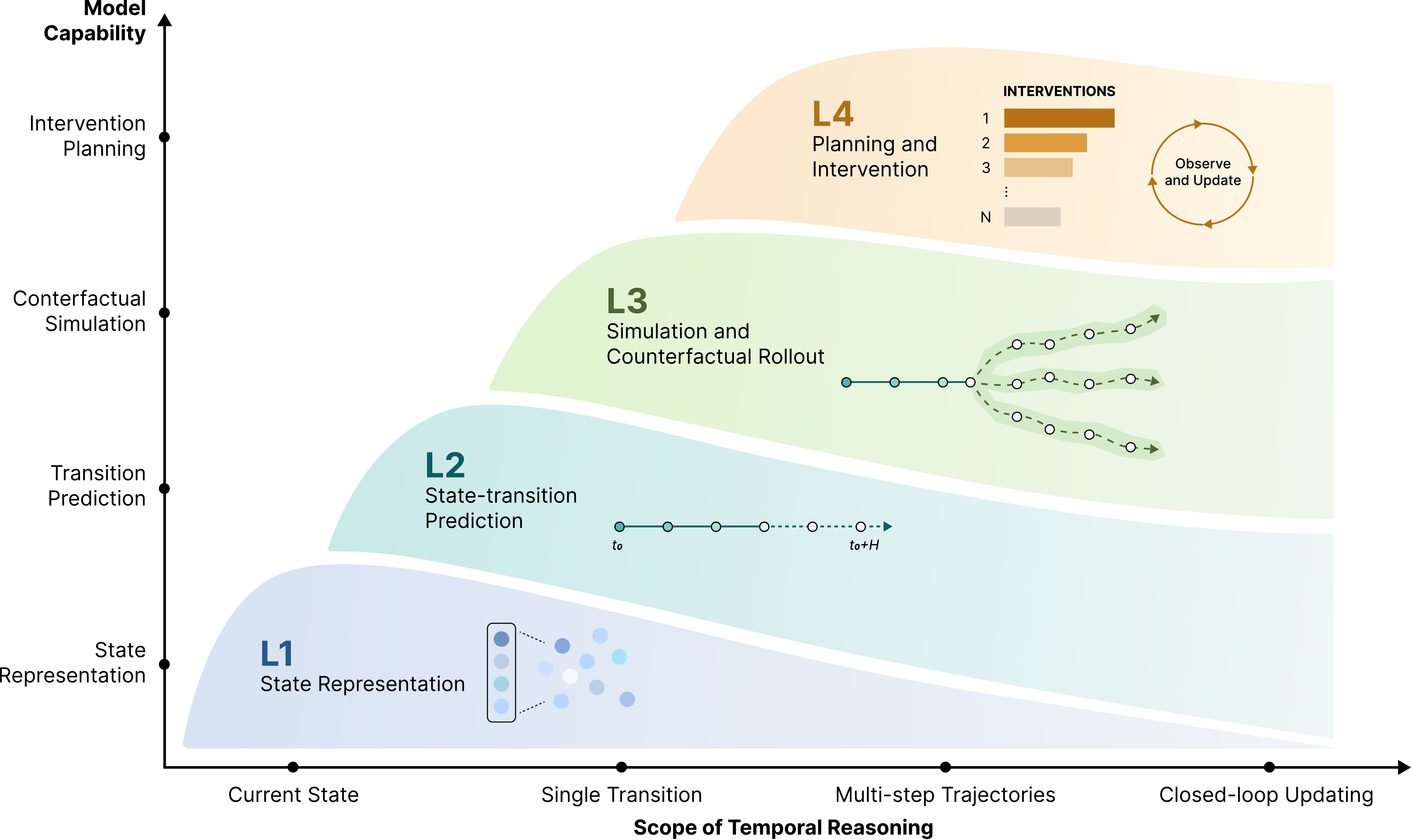}
\caption{\textbf{Capability ladder for Physiological World Models.} Model capability increases vertically from L1 state representation through L2 state-transition prediction and L3 simulation and counterfactual rollout to L4 planning and intervention. The scope of temporal reasoning expands horizontally from the current state to single transitions, multi-step trajectories and closed-loop updating. The nested bands indicate the progressively broader functional scope of each capability level.}\label{fig:capability-ladder}
\end{figure*}

Having reviewed the learning paradigms, we now characterize PWMs according to their functional capabilities through a four-level ladder, summarized in Figure~\ref{fig:capability-ladder}. These levels describe functional capabilities in physiological modelling, rather than degrees of autonomy, clinical readiness, or regulatory maturity. This ladder progresses from L1 state representation to L4 bounded intervention planning; each level builds on the previous one and demands progressively stronger evidence, ranging from unlabelled observational data to causally informative intervention data and prospective validation.

\subsubsection{L1: State Representation}\label{sec:l1-state-representation}

An L1 PWM learns robust and temporally coherent latent representations of HumanState from multimodal physiological observations and interpretable physiological variables~\cite{ref30,ref31}. Inputs may include PPG, ECG, CGM, respiration, EDA, activity, and validated sleep-related measures~\cite{ref16,ref17,ref30,ref31,ref52}. Time-aligned environmental measurements may be used as contextual inputs, but they remain distinct from the latent physiological state~\cite{ref105,ref106}.

The primary output of L1 is a latent state representation or a set of interpretable state estimates, such as recovery, fatigue, or metabolic stability.

L1 can leverage large-scale unlabelled or weakly labelled data, yet multimodal synchronization, sparse annotation, and quality control remain central challenges. Evaluation should therefore emphasize cross-device transfer, generalization across populations, temporal consistency, agreement with established physiological measures, and robustness to noise or missing modalities~\cite{ref33,ref35,ref141}.

\subsubsection{L2: State-Transition Prediction}\label{sec:l2-state-transition-prediction}

An L2 PWM learns an event- or action-conditioned transition distribution from \(\mathbf{s}_{t}\) to \(\mathbf{s}_{t + \Delta t}\). Unlike forecasting based only on prior observations, it predicts how the latent physiological state evolves in response to a specified event, action, or intervention, conditional on the individual's current state and context. L2 therefore relies on HumanState Transition Tokens to learn conditional transition dynamics.

An L2 model may predict latent-state transitions following exercise, a meal, sleep restriction, or acute stress, together with observable outcomes such as heart-rate recovery or postprandial glucose response. It requires clearly defined baseline and response windows, reliable timestamps and event, action, or intervention annotations, sufficient temporal coverage, and contextual information~\cite{ref123,ref140}. Without event or behaviour labels, it may collapse into ordinary forecasting.

Because L2 predicts conditional trajectories rather than isolated point estimates, evaluation should extend beyond pointwise error to assess trajectory shape, response delay, peak response, recovery speed, within-person consistency, and generalization across contexts~\cite{ref140}.

\subsubsection{L3: Simulation and Counterfactual Rollout}\label{sec:l3-simulation-and-counterfactual-rollout}

L3 generates multiple possible future trajectories under alternative events, actions, or interventions. With observational data alone, these outputs should be interpreted as scenario-conditioned rather than counterfactual rollouts. Interpreting these outputs as counterfactual rollouts or estimates of causal effects requires interventional or otherwise causally informative evidence, together with explicit assumptions regarding potential unmeasured confounding, temporal carryover, adherence, and overlap between comparison groups~\cite{ref71,ref79,ref94,ref124}. The central challenge is therefore to determine whether the available evidence supports the proposed alternative-trajectory comparison.

L3 requires high-confidence HumanState Transition Tokens with reliable event timing, complete baseline and response windows, relevant contextual information, and adequate coverage across individuals and settings. Evaluation should assess factual trajectory accuracy, agreement with held-out intervention responses or other empirical validation targets, intervention-ranking accuracy, uncertainty calibration, and recognition of queries outside the model's valid range~\cite{ref87,ref138,ref139,ref162,ref163}. At deployment, population-level distribution-shift monitoring should complement these query-level checks~\cite{ref142}.

\subsubsection{L4: Planning and Intervention}\label{sec:l4-planning-and-intervention}

L4 extends simulation to bounded planning. A planning layer built around the PWM compares candidate interventions under explicit objectives, costs, contraindications, and safety constraints. It ranks or recommends options for human review and updates these estimates as new observations become available~\cite{ref6,ref135,ref136}. L4 requires defined target populations, calibrated uncertainty, interpretable ranking logic, prospective validation, intervention feedback, and evaluation of failure cases and rare adverse responses~\cite{ref65,ref66,ref67,ref145}. Post-deployment distribution-shift monitoring should complement, but not substitute for, these prospective evaluations~\cite{ref142}. Lower-risk domains such as sleep scheduling, exercise recovery planning, and environmental management may provide suitable early testbeds. Higher-risk uses, including medication adjustment or treatment selection, require clinical oversight, ethical review, and clearly defined responsibility boundaries; in these settings, intervention selection and execution should not be autonomous~\cite{ref144}.

Taken together, L1--L4 describe progressively stronger capabilities for modelling physiological state and its transitions. L1 constructs the state representation. L2 predicts event-, action-, or intervention-conditioned transitions within defined contexts. L3 compares alternative future trajectories and, when supported by causally informative evidence, evaluates counterfactual interventions. L4 uses validated transition dynamics to support bounded intervention planning. The associated data and evidence requirements also become progressively more structured: L1 can use large-scale unlabelled or weakly labelled recordings, whereas L2 relies on high-confidence HumanState Transition Tokens, L3 additionally requires causally informative intervention evidence, and L4 further requires intervention feedback and prospective safety evidence.

\section{Data Protocols for Physiological World Models}\label{sec:data-protocols-for-physiological-world-models}

The PWM framework requires temporally aligned physiological observations and state-transition evidence that existing digital-health infrastructure does not routinely provide. Relevant data may combine continuous sensor streams with event annotations, contextual measurements, intervention assignments, and longitudinal outcomes. The value of such datasets lies not in temporal density alone, but in the temporal alignment that enables characterization of how physiological states change within and across individuals. Together, these limitations constrain PWM development. First, despite the widespread adoption of wearable devices and digital-health platforms, these data remain fragmented across devices and institutional systems~\cite{ref41,ref42,ref43,ref53}. Hospital information systems, consumer wearables, and smartphone applications use different communication protocols, sampling schemes, and storage formats. Standards such as FHIR, Open mHealth, and the OMOP common data model improve interoperability, but additional specifications are needed to align dense, event-anchored, multimodal time series~\cite{ref44,ref45,ref54}.

\begin{figure*}[!t]
\centering
\includegraphics[width=\textwidth]{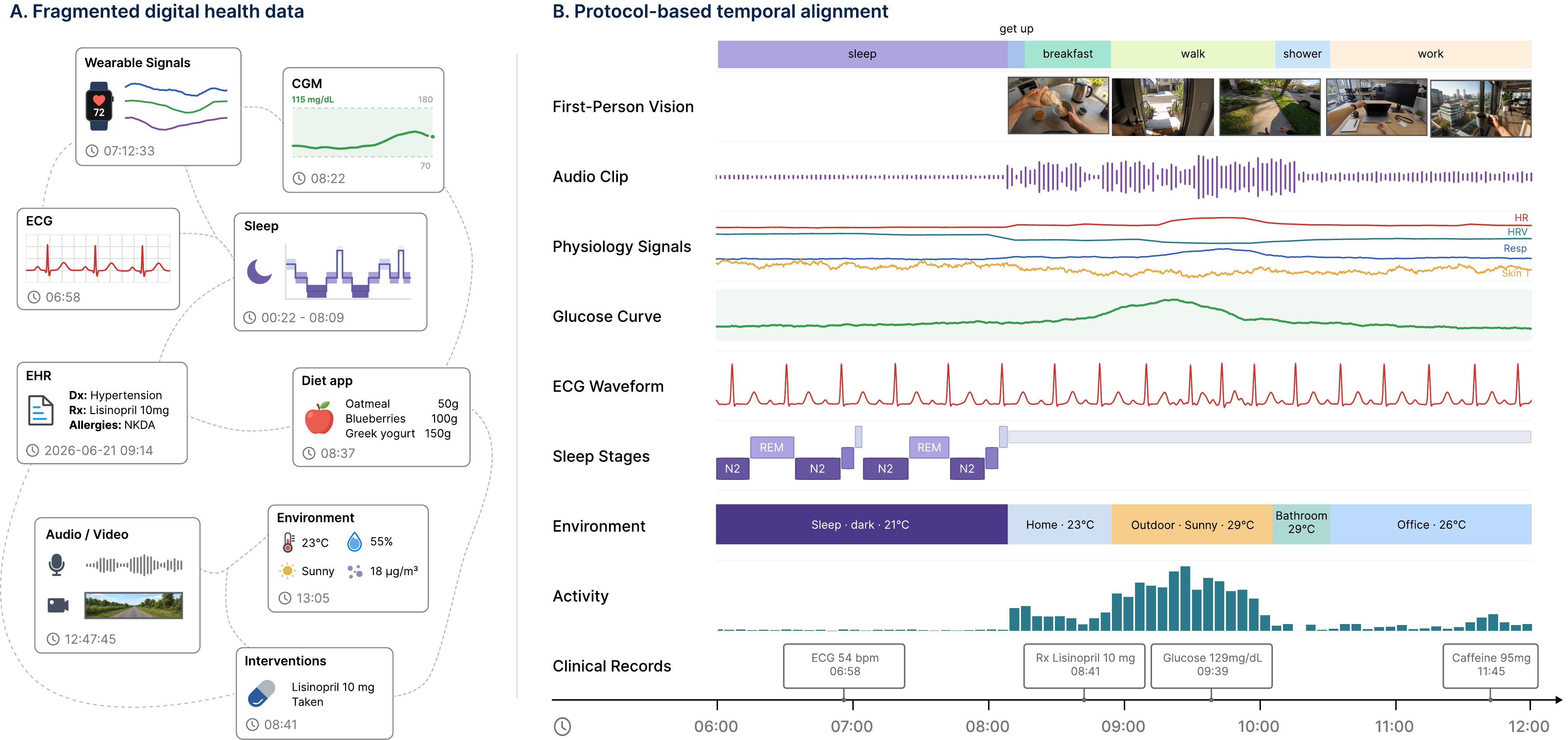}
\caption{\textbf{From fragmented digital-health records to protocol-based temporal alignment.} A, Wearable, metabolic, clinical, behavioural, and environmental data are recorded asynchronously across devices and applications. B, First-person vision, audio, physiological signals, glucose, ECG, sleep, environmental context, activity, and clinical records are organized on a shared event-anchored timeline. Temporal alignment links events and interventions to pre-event baselines and subsequent physiological responses, enabling the construction of comparable HumanState Transition Tokens.}
\label{fig:data-alignment}
\end{figure*}

Figure~\ref{fig:data-alignment} illustrates why interoperability alone is insufficient for PWM development. In fragmented digital-health records, physiological signals and contextual annotations remain distributed across devices, applications, and clinical systems, often with isolated timestamps and poorly resolved cross-modal relationships. Protocol-based temporal alignment instead maps multimodal observations, events, actions, and interventions onto a shared timeline. This common temporal reference defines pre-event baselines and post-event response windows and links physiological changes to their relevant context and observed outcomes, thereby enabling the construction of comparable HumanState Transition Tokens.

A second limitation is that event annotations and intervention records are often sparse, incomplete, or temporally imprecise. A continuous glucose monitor may capture a postprandial glucose excursion with minute-level resolution, while the meal itself may be unlogged or its timing inaccurately self-reported~\cite{ref158}. Ecological momentary assessment can reduce retrospective recall bias and collect contextual information in the moment, but it still relies on participant reporting~\cite{ref46}. Similarly, a wearable ECG device may capture a transient rhythm change without a reliable record of concurrent stressors, activities, or environmental exposures. Such omissions limit the construction of event-anchored transition samples from otherwise extensive physiological recordings.

Third, physiological response windows vary across events and individuals. Exercise, meals, and sleep disruption may produce different response magnitudes, delays, and recovery patterns depending on baseline physiology, disease status, and circadian phase~\cite{ref51,ref52}. Without standardized definitions of pre-event baselines, post-event response windows, outcomes, and data quality, state transitions cannot be compared reliably across individuals, devices, or studies.

Finally, existing datasets remain dominated by passively collected, free-living recordings, often from a single modality. These data are valuable for characterizing baseline distributions and learning latent state representations, but they are insufficient to identify causal intervention effects or to support validation of L3--L4 capabilities~\cite{ref53}. Stronger evidence requires structured event annotation, repeated measurements, assigned interventions where ethical and feasible, and prospective observation of subsequent outcomes.

These limitations motivate a minimum specification for HumanState Transition Tokens and a four-level framework for data acquisition and validation. The protocol levels provide progressively stronger forms of state-transition evidence, but they do not map one-to-one onto PWM capability levels.

\subsection{Minimum Data Protocol}\label{sec:minimum-data-protocol}

To enable HumanState Transition Tokens to serve as standardized and comparable data units across studies, devices, and populations, their constituent fields must be defined with sufficient precision to allow independently collected tokens to be pooled for model training and benchmark evaluation~\cite{ref125,ref127}. We therefore define nine minimum protocol components, each accompanied by examples, minimum recording requirements, and the principal failure modes that arise when those requirements are not met (Table~\ref{tab:minimum-data-protocol}).

\begin{sidewaystable*}[p]
\centering
\caption{Minimum data requirements and associated failure modes for HumanState Transition Tokens.}
\label{tab:minimum-data-protocol}

\fontsize{9.5}{11}\selectfont
\setlength{\tabcolsep}{3pt}
\renewcommand{\arraystretch}{1.02}

\begin{tabularx}{\textheight}{@{}
  >{\raggedright\arraybackslash}p{0.14\textheight}
  >{\raggedright\arraybackslash}X
  >{\raggedright\arraybackslash}X
  >{\raggedright\arraybackslash}p{0.28\textheight}
@{}}

\arrayrulecolor{FulliveInk}
\toprule

\textbf{Component} &
\textbf{Examples} &
\textbf{Required precision} &
\textbf{Failure modes} \\
\midrule

Time Synchronization &
All sensor streams aligned to a common UTC-anchored timeline &
Clock drift and timestamp accuracy documented; synchronization accuracy matched to the modality and expected physiological response timescale &
Temporal misalignment between event markers and physiological response windows; spurious lead-lag relationships \\
\densetablerowrule

Sensor Modalities &
PPG, ECG, CGM, actigraphy, respiration, EDA, ambient temperature, light, audio level &
Device model and firmware version documented; sampling rate and filter settings specified &
Inability to pool data across sensor generations; unmodelled differences in signal fidelity \\
\densetablerowrule

Event Ontology &
Noise exposure, meeting (type, intensity), ambient temperature change, light exposure change, social interaction, air quality &
Event category drawn from a standardized, hierarchical taxonomy of external events and exposures; event onset or exposure interval recorded to the required accuracy &
Event ambiguity across studies; inability to compare ``commute'' responses when commute mode, duration, and peak stressor timing differ; unmeasured external events that explain physiological fluctuations misattributed to spontaneous variation \\
\densetablerowrule

Action Log &
Eating (food type and approximate quantity), exercise (type, intensity, duration), medication (drug, dose), caffeine, alcohol, screen use, sleep onset/offset &
Structured, timestamped log entries; smartphone-based or voice-assisted entry used where feasible to minimize reliance on retrospective recall &
Recall bias; incomplete or incorrect action labelling that contaminates event-conditioned learning; unrecorded actions that leave physiological traces in the data but are absent from the log, causing the model to attribute their effects to other variables \\
\densetablerowrule

Context Variables &
Age, sex, chronic conditions, medications, fitness level, chronotype, baseline HR/HRV/CGM setpoints from a pre-specified observation period appropriate to the modality and task; location, activity state, prior sleep duration and quality, circadian phase, ambient conditions &
Standardized demographic and clinical coding; location from GPS or beacon at a spatial resolution appropriate to the contextual exposure of interest; sleep duration and quality derived from wearable sleep staging validated against reference polysomnography~\cite{ref164}; circadian phase estimated from activity and light exposure &
Baseline physiology recorded during an unrepresentative period (e.g., acute illness, travel); undocumented conditions or unreported medications that alter response dynamics; omitted situational moderators (e.g., location, prior sleep debt) that explain response heterogeneity across otherwise identical tokens \\
\densetablerowrule

Intervention Record &
Dietary regimen, exercise prescription, sleep schedule adjustment, medication change &
Intervention type, dose or intensity, adherence metrics, start and end timestamps &
Inability to distinguish intervention effects from spontaneous physiological variation; unmeasured non-adherence \\
\densetablerowrule

Baseline Observation Window &
Continuous physiological observations over a pre-specified window preceding the focal event, action, or intervention, with a task-specific missingness threshold &
Sufficient to compute baseline HR, HRV, glucose, activity level, and stress indices &
Unstable or unrepresentative baseline contaminating the estimated state-transition magnitude \\
\densetablerowrule

Physiological Response Window &
Continuous physiological observations over a pre-specified window following the focal event, action, or intervention &
Duration matched to the expected physiological response timescale of the event category; missingness \textless{} 20\% &
Truncated response window; missed delayed effects; inability to characterize response dynamics \\
\densetablerowrule

Outcome and Data Quality &
Quantifiable physiological change (e.g., postprandial glucose AUC, HRV recovery slope, sleep efficiency) plus documented quality indicators covering completeness, temporal alignment, sensor quality, annotation reliability, and overall confidence &
Outcome measure defined a priori for each event category; quality score computed from the five high-confidence token dimensions &
Uninterpretable or misleading transition samples entering the training corpus; inflated or unstable model performance estimates \\

\arrayrulecolor{FulliveInk}
\bottomrule

\end{tabularx}
\end{sidewaystable*}

This Minimum Data Protocol serves two purposes. First, it provides a concrete specification that the data producers, such as researchers, device manufacturers, and clinical programmes, can use to design and evaluate data collection work~\cite{ref126,ref127}. Second, it defines the minimum quality requirements that determine whether a token is suitable for training or evaluation at a given capability level~\cite{ref125,ref126}.

\subsection{Data Acquisition and Validation Levels}\label{sec:data-acquisition-and-validation-levels}

Drawing on fit-for-purpose evaluation principles for connected sensor technologies~\cite{ref125,ref126} and FAIR data-stewardship principles~\cite{ref127}, we propose four PWM-specific levels of data acquisition and validation, designated P1 through P4. These levels differ in the precision of event annotation, the degree of intervention control, the availability of prospective feedback, and the strength of the state-transition evidence they provide. Their relationship to the PWM capability levels is enabling rather than one-to-one: a capability may depend on evidence from multiple protocol levels, and no protocol level alone guarantees that the corresponding model capability has been achieved.

\paragraph{P1 --- Passive Free-Living Recording} At this level, data are collected passively during participants' normal daily lives without assigned activities or experimental interventions. The primary objective is to characterize baseline physiological distributions and their natural temporal variability. Relevant measurements may include heart rate, HRV, glucose, activity, estimated sleep stages, and their circadian and day-to-day patterns.

P1 data are often noisy. Events are often sparsely labelled or inferred from sensor patterns, while contextual information and intervention records may be incomplete. P1 therefore provides the observational foundation for self-supervised or weakly supervised learning of latent HumanState representations, but it does not necessarily produce high-confidence HumanState Transition Tokens. Large-scale cohorts with wearable components, such as the UK Biobank accelerometry study~\cite{ref47} and the All of Us Research Program~\cite{ref48}, illustrate important elements of P1 data collection at population scale.

\paragraph{P2 --- Structured Event and Action Annotation Protocol} At P2, participants remain in free-living settings but follow structured event- and action-logging and measurement procedures. Meals, exercise sessions, sleep timing, and other relevant events are recorded using prespecified fields and reliable timestamps. Examples include meal logs supported by photographs, timestamped exercise sessions with intensity ratings, sleep diaries, and in-the-moment self-documentation or ecological momentary assessment~\cite{ref46}. P2 does not require the assignment of an intervention condition.

The objective of P2 is to construct comparable, event-anchored transition samples. P2 therefore can provide structured observational data that support L2 event- and action-conditioned state-transition learning.

\paragraph{P3 --- Controlled Intervention Studies} At P3, specific behavioural, environmental, or clinical interventions are deliberately assigned, and their physiological effects are measured under controlled or semi-controlled conditions. Examples include randomized crossover studies comparing prescribed exercise with habitual activity, studies comparing controlled dietary conditions, and protocols that systematically advance or delay sleep timing~\cite{ref129}.

Randomized crossover and repeated-measures designs can provide stronger evidence for within-person intervention effects, provided that randomization, adherence, period effects, carryover effects, and relevant time-varying confounders are appropriately addressed~\cite{ref92,ref93}. These protocols can produce high-confidence transition tokens containing intervention assignments, pre-intervention states, contextual variables, post-intervention trajectories, outcomes, and data-quality information.

P3 provides an important empirical basis for L3 intervention comparison and counterfactual evaluation. Nevertheless, an outcome observed under a different condition or at another time should not be treated as ground truth for the unobserved individual-level counterfactual.

\paragraph{P4 --- Adaptive Prospective and Shadow-Mode Validation} P4 targets event--context combinations and physiological conditions that remain sparsely represented after P1--P3. At this level, model predictions or intervention rankings are generated prospectively before the corresponding outcomes are observed; in shadow mode, they do not influence participant behaviour or clinical care. Subsequent real-world observations are then used to evaluate predictive accuracy, calibration, failure modes, and robustness under distributional shift.

Model uncertainty and observed failures may guide additional data collection or ethically bounded intervention studies, directing empirical data collection towards poorly characterized regions of the state-transition space~\cite{ref130}.

Clearly labelled synthetic trajectories may complement P4 by supporting hypothesis generation and the exploration of rare scenarios~\cite{ref49}, but cannot replace empirical observations, causal evidence, prospective validation or evidence of intervention safety.

P4 supports the validation requirements of L3 and L4 rather than independently enabling L4 planning. Any progression towards L4 additionally requires defined target populations, explicit safety constraints, prospective intervention evidence, failure-case analysis, long-term follow-up, and appropriate human or clinical oversight.

These levels are complementary rather than strictly cumulative, and a study may combine elements from several levels. Progress from L1 to L4 therefore requires not only larger datasets, but also increasingly structured, causally informative, and prospectively validated evidence. Standardized and reusable data should follow FAIR data-stewardship principles~\cite{ref127}, whereas measurement and clinical validation should follow fit-for-purpose evaluation frameworks~\cite{ref125,ref126}. Causally informative and prospective evidence additionally requires appropriate intervention designs and prospective clinical evaluation~\cite{ref67,ref71,ref79}. Progress in physiological world modelling will depend not only on advances in model architecture but also on rigorous data acquisition, well-designed intervention studies, and prospective validation.

\section{Benchmarking and Applications}\label{sec:benchmarking-and-applications}

Widely used clinical time-series and biomedical-signal benchmarks evaluate tasks such as in-hospital mortality prediction, physiological decompensation, length-of-stay forecasting, phenotype classification, diagnosis, and digital-biomarker classification~\cite{ref155,ref156}. Although these tasks remain valuable, they do not establish whether a model can represent an integrated HumanState, predict event-conditioned transitions, simulate trajectories under alternative interventions, or recognize when its predictions are unreliable.

We therefore define six complementary benchmark tasks. Tasks 1--5 evaluate progressively stronger modelling capabilities, whereas Task 6 assesses reliability across all capability levels. Together, these tasks connect the PWM capability hierarchy, data protocols, and potential applications.

\subsection{Benchmark Framework}\label{sec:benchmark-framework}

\begin{table*}[t]
\centering
\caption{Benchmark tasks for Physiological World Models.}
\label{tab:benchmark-tasks}

\fontsize{8.5}{10}\selectfont
\setlength{\tabcolsep}{4pt}
\renewcommand{\arraystretch}{1.06}

\begin{tabularx}{\textwidth}{@{}
  >{\raggedright\arraybackslash}p{0.18\textwidth}
  >{\raggedright\arraybackslash}X
  >{\raggedright\arraybackslash}X
  >{\raggedright\arraybackslash}p{0.18\textwidth}
@{}}

\arrayrulecolor{FulliveInk}
\toprule
\textbf{Task} &
\textbf{Evaluation objective} &
\textbf{Representative metrics} &
\textbf{Capability and evidence requirements} \\
\midrule

\textbf{T1:} HumanState Representation and Estimation &
Determine whether the model constructs a physiologically meaningful and transferable HumanState representation &
Agreement with validated physiological measures; cross-device and cross-population transfer; robustness to noise and missing modalities &
L1; primarily P1 \\
\tablerowrule

\textbf{T2:} Multi-horizon State-Transition Forecasting &
Predict how HumanState evolves across multiple timescales following a specified event, action, or intervention, conditional on the current HumanState and Context &
Trajectory error; effective forecasting horizon; prediction-interval coverage; calibration &
L2; primarily P2 \\
\tablerowrule

\textbf{T3:} Event-Conditioned and Individualized Response Prediction &
Determine whether the model captures person-specific response patterns and heterogeneity following a given event, action, or intervention &
Response direction; peak amplitude; time to peak; recovery rate; personalization gain &
L2; P2--P3 \\
\tablerowrule

\textbf{T4:} Alternative Intervention Simulation &
Compare plausible trajectories under alternative interventions &
Held-out intervention error; intervention-ranking accuracy; cross-condition generalization error; uncertainty calibration &
L3; primarily P3 \\
\tablerowrule

\textbf{T5:} Bounded Planning and Steerability &
Rank intervention options under explicit objectives, costs, contraindications, and safety constraints &
Prospective outcome improvement; intervention efficiency; ranking accuracy; safety violations &
L4; P3--P4 \\
\tablerowrule

\textbf{T6:} Reliability under Distribution Shift &
Determine whether the model recognizes when its predictions become unreliable &
Robustness gap; calibration error; uncertainty--error association; abstention performance; subgroup failure analysis &
Cross-cutting across L1--L4; P1--P4, with external or prospective evaluation under relevant distribution shifts \\

\arrayrulecolor{FulliveInk}
\bottomrule
\end{tabularx}
\end{table*}

Because latent HumanState is not directly observable, T1 should combine several forms of evidence rather than rely on a single latent-space metric. These may include agreement with validated measures, downstream transfer, robustness to missing modalities, and generalization across devices and populations~\cite{ref30,ref33,ref34,ref35,ref141}.

T2 evaluates whether the model can forecast HumanState trajectories conditioned on events, actions, or intervention across multiple prediction horizons. T3 provides a stricter test of whether the model captures person-specific response heterogeneity and improves prediction through individualization. Forecasting based only on prior observations may serve as a baseline, but should not be treated as evidence of L2 Physiological World Model capability.

T4 requires a clear distinction between scenario simulation and counterfactual interpretation, with the latter requiring data from randomized, crossover, or other causally informative study designs. The results observed under the held-out intervention conditions can provide empirical validation targets but do not constitute ground truth for the unobserved individual-level counterfactual~\cite{ref79,ref94}.

T5 should initially be evaluated prospectively in shadow mode and under explicit safety constraints~\cite{ref67,ref150}. Evaluation across clinical contexts should precede broader deployment~\cite{ref151}. T6 applies to every preceding task and should function as a reliability gate for higher-risk applications, with explicit assessment of uncertainty, robustness and harmful distribution shifts~\cite{ref80,ref81,ref139,ref147,ref149}.

\subsection{Application-Specific Benchmark Profiles}\label{sec:application-specific-benchmark-profiles}

The six benchmark tasks defined above serve complementary evaluation purposes. Translating task performance into application readiness requires a principled mapping between each use case and the appropriate combination of tasks, capability levels, and evidence standards. Table~\ref{tab:application-horizons} summarizes this mapping for representative applications across near-, mid-, and long-term horizons. These horizons indicate an expected progression in capability, evidence, and governance requirements rather than fixed deployment timelines.

\begin{table*}[t]
\centering
\caption{Application horizons and indicative benchmark profiles.}
\label{tab:application-horizons}

\fontsize{8.5}{10}\selectfont
\setlength{\tabcolsep}{4pt}
\renewcommand{\arraystretch}{1.08}

\begin{tabularx}{\textwidth}{@{}
  >{\raggedright\arraybackslash}p{0.13\textwidth}
  >{\raggedright\arraybackslash}p{0.20\textwidth}
  >{\raggedright\arraybackslash}p{0.12\textwidth}
  >{\raggedright\arraybackslash}p{0.12\textwidth}
  >{\raggedright\arraybackslash}X
@{}}

\arrayrulecolor{FulliveInk}
\toprule

\textbf{Application horizon} &
\textbf{Representative applications} &
\textbf{Required tasks} &
\textbf{Capability level} &
\textbf{Evidence and validation requirements} \\
\midrule

\textbf{Near-term} &
Sleep and fatigue monitoring; exercise recovery forecasting; metabolic stability monitoring; stress-related recovery &
T1, T2, T3 and T6 &
Primarily L1--L2 &
P1--P2 data; external validation; personalization tests; calibration and applicability boundaries \\
\tablerowrule

\textbf{Mid-term} &
Meal- and caffeine-timing guidance; sleep scheduling; exercise planning; work--rest adjustment &
T1--T4 and T6 &
Primarily L2--L3 &
P2--P3 data; repeated or N-of-1 observations; crossover interventions; intervention-ranking validation \\
\tablerowrule

\textbf{Long-term} &
Chronic disease trajectories; adaptive rehabilitation planning; clinical-trial enrichment; bounded clinical decision support &
T1--T6 &
Primarily L3--L4 &
P3--P4 evidence; prospective validation; safety gates; clinical oversight; regulatory and accountability requirements \\

\arrayrulecolor{FulliveInk}
\bottomrule

\end{tabularx}
\end{table*}

No single benchmark task or capability level is sufficient to establish application readiness. For example, personalized recovery management requires reliable state representation, multi-horizon prediction, individualized response modelling, and uncertainty calibration. Behavioural intervention design additionally requires credible comparison of alternative trajectories. Higher-risk clinical applications require the complete benchmark profile, including prospective planning evaluation and reliability under distribution shift.

\subsubsection{Near-term: Personalized Health and Recovery}\label{sec:near-term-personalized-health-and-recovery}

Near-term PWM applications are likely to focus on lower-risk personal health and recovery management. By integrating sleep-related measures, HRV, activity load, body-temperature rhythms, dietary events, and environmental context, a PWM could estimate recovery trajectories following sleep restriction, stress exposure, exercise, or metabolic perturbation~\cite{ref52,ref56,ref57,ref58,ref59,ref60,ref88}.

Potential users include athletes, shift workers, healthcare workers, and other populations exposed to demanding or irregular schedules. These applications should provide probabilistic estimates with explicit uncertainty and applicability boundaries rather than deterministic behavioural instructions.

\subsubsection{Mid-term: Behavioural Intervention Design}\label{sec:mid-term-behavioural-intervention-design}

Mid-term applications may focus on the comparison of behavioural interventions, including changes in meal timing, caffeine use, sleep environment, exercise scheduling, and work--rest patterns~\cite{ref52,ref58,ref61,ref62}. The aim is not to identify a universally optimal behaviour, but to compare plausible physiological trajectories under alternative choices while accounting for the individual's current state, baseline, history, and context.

In the absence of causally informative intervention data, these comparisons should be described as scenario simulations; causal claims require data from randomized N-of-1 or crossover studies, or other study designs that support causal identification~\cite{ref96,ref128}.

\subsubsection{Long-term: Higher-risk Clinical Decision Support}\label{sec:long-term-higher-risk-clinical-decision-support}

Longer-term applications may include chronic disease trajectory modelling, adaptive rehabilitation planning, metabolic disease management, treatment-toxicity monitoring, clinical-trial enrichment, and in silico patient-response simulation~\cite{ref9,ref10,ref63,ref64,ref75,ref76,ref137}. In these settings, PWMs should support trajectory exploration, hypothesis generation, and clinician-in-the-loop decision-making rather than autonomous treatment selection.

Risk estimates may be derived from predicted trajectories, but the defining capability remains the modelling and comparison of physiological state transitions. Simulated trajectories should not be treated as replacements for clinical trials or as sufficient evidence for independent treatment decisions.

Higher-risk applications require prospective validation, safety-gated evaluation, calibration assessment, failure-case analysis, and clearly defined target populations and accountability boundaries~\cite{ref65,ref66,ref67,ref165}. Applications involving treatment selection, medication adjustment, or intervention planning additionally require regulatory oversight and continuing clinical supervision.

Across these horizons, evidence requirements progress from reliable state representation and transition prediction to causally informative intervention evaluation, bounded planning, prospective validation, and robust recognition of model limitations.

\section{Challenges, Open Questions, and Scope Boundaries}\label{sec:challenges-open-questions-and-scope-boundaries}

The central challenge for PWMs is not simply increasing data volume. It is establishing reliable evidence for physiological state representation, state-transition prediction, and intervention comparison from heterogeneous, biased, and incomplete real-world records. Six areas require particular attention.

\subsection{Data Heterogeneity and Evidence Quality}\label{sec:data-heterogeneity-and-evidence-quality}

The data bottleneck differs across capability levels. L1 requires large-scale physiological observations that are sufficiently harmonized for representation learning, whereas L2--L4 increasingly depend on well-aligned and verifiable HumanState Transition Tokens. Wearables, CGMs, smartphones, sleep-monitoring devices, and clinical records differ in sampling frequency, device algorithms, missingness patterns, and annotation quality. Motion artefacts, non-wear periods, event ambiguity, and population bias can make measurements of the same physiological variable difficult to compare across devices and populations~\cite{ref68,ref69,ref97,ref99,ref104}.

Future data infrastructure should therefore prioritize interoperability, provenance, temporal alignment, and the quality of state-transition evidence rather than recording duration alone. Device characteristics, preprocessing procedures, event definitions, missingness patterns, and data-quality indicators should be documented in forms that allow observations and transition tokens to be compared across studies and populations.

\subsection{Temporal Confounding and Causality}\label{sec:temporal-confounding-and-causality}

Real-world events rarely occur in isolation. Sleep loss, stress, meals, caffeine, exercise, medication use, and environmental exposure often co-occur and affect physiological state with different delays. From observational trajectories alone, it is difficult to distinguish genuine intervention effects from spontaneous recovery and confounding by co-occurring exposures.

Repeated within-person measurements and randomized N-of-1 or crossover designs can strengthen evidence about intervention effects~\cite{ref71}. Negative controls can help detect residual confounding and bias in observational analyses~\cite{ref70}. Causal sensitivity analyses should assess how conclusions change under plausible levels of unmeasured confounding~\cite{ref159}. Because controlled study conditions differ from everyday life, evidence from these designs should be complemented by evaluation in free-living settings, where prediction reliability can be monitored across devices, populations, and timescales.

\subsection{Personalization and Continual Adaptation}\label{sec:personalization-and-continual-adaptation}

An individual's physiological baseline is not fixed. Ageing, disease progression, training load, medication, seasonality, infection, and changes in daily routine can alter how the same event translates into a physiological response. PWMs therefore require continual calibration, but continual adaptation can also introduce model drift and safety risks~\cite{ref37}.

Future systems should separate state updating, individual calibration, model-parameter updating, and the generation of high-risk recommendations. These processes require separate validation and monitoring procedures. Continual adaptation should also include model-version tracking, rollback mechanisms, and checks for catastrophic forgetting, calibration drift, and performance degradation across population subgroups~\cite{ref142,ref149,ref150,ref152,ref157}.

\subsection{Safety, Interpretability, and Trust}\label{sec:safety-interpretability-and-trust}

PWM predictions and model-supported recommendations should not be presented as certain or definitive. They should be accompanied by appropriately calibrated uncertainty estimates, applicability boundaries, and clearly defined abstention or escalation criteria. Users need to understand the basis of lower-risk predictions, while clinicians require sufficient information to accept, reject, or modify model-supported recommendations.

Interpretability analyses should relate latent representations of HumanState to validated physiological measures and clinically meaningful outcomes. Feature attribution alone should not be interpreted as evidence of a causal physiological mechanism. Explanations should also remain sufficiently stable across small changes in input, device, and model version to support meaningful review~\cite{ref166,ref167,ref168,ref169}.

In medical settings, clinician-in-the-loop oversight, prospective validation, transparent reporting, and predefined responsibility boundaries are prerequisites for deployment~\cite{ref65,ref66,ref67}. Safety governance should address overconfidence, rare adverse responses, distributional shift, inappropriate extrapolation, and failures that may not be captured by average predictive performance.

\subsection{Privacy, Governance, and Equity}\label{sec:privacy-governance-and-equity}

PWMs may combine physiological signals with behaviour, location, sleep, diet, environmental exposure, and, where explicitly included, audio- or image-derived contextual information. Combining these sources can reveal sensitive patterns that would not be apparent in any single data stream, creating privacy and governance risks beyond those associated with any individual data source considered in isolation~\cite{ref170}.

Privacy-by-design may include federated or local learning combined with encryption, differential privacy, access control, and traceability~\cite{ref72}. Governance frameworks should additionally specify data-minimization requirements and auditable data-use policies, and should record data provenance, permitted uses, device characteristics, processing histories, access records, and whether a transition sample was empirically observed or synthetically generated.

For synthetic data, provenance records should include the generating model, model version, conditioning variables, and intended use. They should also indicate whether a sample was used for training or evaluation. Synthetic samples should remain distinguishable from empirical observations throughout the model lifecycle and should not be treated as intervention or safety evidence.

Equity is also essential. PWM development should not privilege users of high-end devices or populations already well represented in digital health datasets. Validation across devices, ages, socioeconomic groups, care settings, and, where relevant to optical sensing, skin tones should form part of the core evaluation framework~\cite{ref73,ref74,ref171}. Performance differences should be reported alongside aggregate results rather than obscured by population-level averages.

\subsection{Scope Boundaries and Relation to Digital Twins}\label{sec:scope-boundaries-and-relation-to-digital-twins}

Digital twins and PWMs are overlapping but distinct. Digital twins generally emphasize an individual-specific virtual representation, continuous state synchronization, and closed-loop interaction with the represented system. PWMs emphasize learnable state-transition dynamics, conditional simulation, and intervention comparison~\cite{ref75,ref76}. It is therefore too strong to describe digital twins simply as a subset of PWMs or to assume that a personalized PWM automatically constitutes a complete digital twin.

A mature PWM could serve as the core dynamics module of some human digital twins. A complete digital twin would additionally require reliable identity mapping, continuous data synchronization, interface design, governance, workflow integration, and clearly defined accountability boundaries.

Early PWMs need not be developed as full digital twins. They may begin with lower-risk transition-modelling tasks such as sleep recovery, metabolic stability, exercise recovery, and rehabilitation trajectories. Integration into more complete personalized simulation or decision-support systems should occur only after the relevant capabilities have been prospectively validated.

Taken together, these challenges indicate that progress from state representation to intervention planning should be gated not only by model performance, but also by stronger empirical evidence, prospective validation, transparent governance, and clearly defined limits of use.

\section{Outlook}\label{sec:outlook}

The near-term value of Physiological World Models does not lie in simulating the human body in its entirety. That goal remains distant and may be unnecessary for macro-level physiological applications. Instead, PWMs can provide an event-conditioned framework that organizes heterogeneous physiological and contextual data around state transitions rather than according to particular devices or institutional sources.

Realizing this potential requires a staged roadmap across models, data, and sensing. At the model level, development should proceed from robust multimodal state representation to event-conditioned transition prediction, simulation of alternative trajectories, and bounded planning. Each advance should be gated by calibrated uncertainty, clearly defined applicability limits, and progressively stronger empirical validation. In parallel, longitudinal and event-anchored datasets must span diverse populations, devices, event types, interventions, contexts, and timescales. Standardized protocols, improved sensing accuracy and long-term wearability, cross-device interoperability, and prospective shadow-mode evaluation will be needed to establish whether transition predictions remain reliable across devices and populations.

Together, these advances could form a shared infrastructure for computational physiology, with PWMs serving as a unifying computational layer that connects physiological observation, state-transition modelling, and evidence-based evaluation. Such an infrastructure could move health AI beyond describing current states towards estimating conditional future trajectories and, where supported by causally informative evidence, comparing alternative courses of action.

\backmatter

\balance
\bibliography{manuscript-references}

\end{document}